\documentclass[sn-basic]{sn-jnl}

\usepackage{graphicx}
\usepackage{multirow}
\usepackage{makecell}
\usepackage{pifont}
\usepackage{array}
\usepackage{bm}
\usepackage{subfigure}

\jyear{2021}%

\theoremstyle{thmstyleone}%
\theoremstyle{thmstyletwo}%

\theoremstyle{thmstylethree}%

\begin{document}

\title[SSR-TA: Sequence to Sequence based expert recurrent recommendation...]{SSR-TA: Sequence to Sequence based expert recurrent recommendation for ticket automation}


\author[1]{\fnm{Chenhan} \sur{Cao}}\email{1220044906@njupt.edu.cn}

\author[1]{\fnm{Xiaoyu} \sur{Fang}}\email{1220044932@njupt.edu.cn}

\author[1]{\fnm{Bingqing} \sur{Luo}}\email{luobq@njupt.edu.cn}

\author*[1]{\fnm{Bin} \sur{Xia}}\email{bxia@njupt.edu.cn}

\affil*[1]{\orgname{Jiangsu Key Laboratory of Big Data Security and Intelligent Processing, Nanjing University of Posts and Telecommunications}, \orgaddress{\city{Nanjing}, \country{China}}}




\abstract{The ticket automation provides crucial support for the normal operation of IT software systems. An essential task of ticket automation is to assign experts to solve upcoming tickets. However, facing thousands of tickets, inappropriate assignments will make tickets transfer frequently among experts, which causes time delays and wasted resources. Effectively and efficiently finding an appropriate expert in fewer steps is vital to ticket automation. In this paper, we proposed a sequence to sequence based translation model combined with a recurrent recommendation network to recommend appropriate experts for tickets. The sequence to sequence model transforms the ticket description into the corresponding resolution for capturing the potential and useful features of representing tickets. The recurrent recommendation network recommends the appropriate expert based on the assumption that the previous expert in the recommendation sequence cannot solve the expert. To evaluate the performance, we conducted experiments to compare several baselines with SSR-TA on two real-world datasets, and the experimental results show that our proposed model outperforms the baselines. The comparative experiment results also show that SSR-TA has a better performance of expert recommendations for user-generated tickets.}



\keywords{ticket automation, seq2seq, recurrent recommendation, description translation}

\maketitle

\section{Introduction}\label{sec1}
\begin{figure}[ht] 
\centering 
\includegraphics[width=8cm]{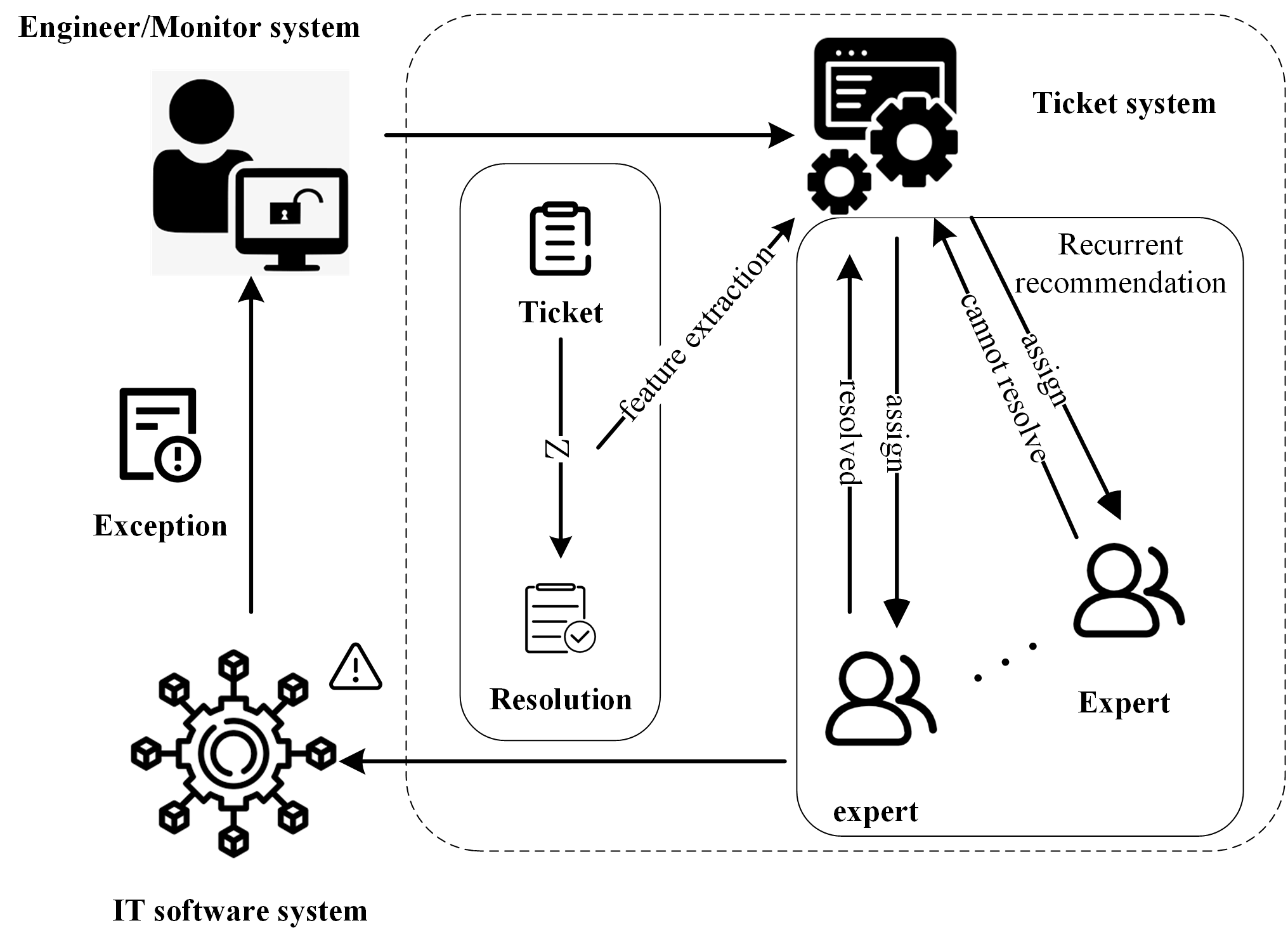} 
\caption{\centering The workflow of the ticket automation} 
\label{fig.ticketsystem} 
\end{figure}

In an IT software system, maintaining the normal operation is the essential requirement to provide high-quality services. In practice, the relationship among components of the IT software system is complicated. Once an exception occurs in a component, the exception may spread rapidly to other components and generate many concurrent warnings, causing the system unable to provide services. For example, IBM Cloud suffered two outages within five days in 2021~\cite{IBM}. The services were hit in the UK, the US, Sydney, Tokyo, and more for several hours due to incorrect routing settings by the external network provider. However, manually assigning an expert to address the exception is time-consuming and expensive. In order to catch exceptions and solve problems in time, the ticket automation is widely used in IT software systems. Fig.~\ref{fig.ticketsystem} illustrates the workflow of the ticket automation. In detail, when an exception is detected by the engineer/monitor system, a ticket containing the exception description and the corresponding system information will be submitted to the ticket system. Further, the ticket system assigns an expert to solve the ticket according to the submitted information (i.e., the description and the system information). If the current expert cannot solve the ticket, another expert will be assigned by the ticket system until the ticket is solved. However, the inappropriate assignment (i.e., the assigned expert cannot solve the ticket) will cause serious problems such as time delays, wasted resources, and service failures. It is crucial to assign an appropriate expert to solve the ticket in time.

Many efforts have been made by researchers to improve the performance of ticket automation at different stages. For the ticket generation, the previous works focused on generating tickets automatically~\cite{palacios2019big} and reducing ticket submitting delays~\cite{gupta2018reducing}. To represent ticket description, the approaches based on the service catalog mapping~\cite{shimpi2014problem} and the entity name extraction~\cite{han2018towards} were introduced. Additionally, several methods were proposed for the ticket classification, including the entity recognition for ticket description~\cite{potharaju2013juggling}, the similarity based ticket clustering~\cite{xu2018signature}, and the hierarchical multi-label classification~\cite{zeng2017knowledge}. Furthermore, researchers proposed two types of ticket recommendation methods, including resolution and expert recommendation. The resolution recommendation is the methods based on the ticket description similarity~\cite{zhou2017star} or the hierarchical multi-armed bandit~\cite{wang2018aistar} recommend a corresponding resolution for the ticket. The expert recommendation is proposed to find an appropriate expert to solve the ticket based on the ticket description and the ticket transfers among experts~\cite{han2020deeprouting}. However, the recommendation method based on similarity may cause that the similar experts who cannot solve the ticket will be recommended in the same list due to the inappropriate representations of tickets. 

To overcome the aforementioned problems, we propose a Sequence to Sequence (seq2seq) based model SSR-TA which is combined with a recurrent recommendation network to recommend appropriate experts for the ticket automation. For training SSR-TA, the description of solved tickets is translated into the corresponding resolution, where the potential features of ticket description and resolution are captured to improve the representation of the ticket description. The expert recommendation will be generated based on the effective representation of the ticket description. Because the traditional Top-N recommendation generates expert sequences based on the similarity, which may cause that if the current expert cannot solve the ticket, other similar experts in the sequence cannot solve the ticket either. By introducing an attention mechanism, SSR-TA recurrently recommends the next appropriate expert based on the assumption that the previous expert in the recommended sequence failed to solve the ticket. The main contributions of this paper are summarized as follows:

\begin{enumerate}
    \item To the best of our knowledge, SSR-TA is the first seq2seq based model that combined the features from description and resolution to assist the expert recommendation, and the model can be used for ticket resolution generation. 
    
    \item We propose a recurrent recommendation network adopting an attention mechanism to generate the expert recommendation sequences, while the generation is based on the assumption that the previous expert in the sequence failed to solve tickets, the results show that the recurrent recommendation network improves the ranking of the true resolver in the recommendation sequence. 
    
    \item We conduct experiments to discuss the effectiveness of description translation and the recurrent recommendation network, and compare the performance of several baselines with SSR-TA.   
\end{enumerate}

The rest of the paper is organized as follows: Section 2 summarizes research related to ticket automation. Section 3 describes the details of SSR-TA. Section 4 conducts experimental comparisons and results analysis. Finally, Section 5 concludes the paper.

\section{Related Work}\label{sec2}
This section introduces the related research on the ticket automation and the application of seq2seq technology. In addition, the similarities and differences between these approaches and ours are presented briefly.

\subsection{Ticket automation}\label{subsec1}
\textbf{Ticket data analysis:} For automatic management of cellular networks, Palacios et al.~\cite{palacios2019big} proposed a big data-empowered framework using the anomaly detection and root cause analysis tools to indentify the network state and automatically generate trouble tickets. To reduce the time of interaction during resolving a ticket, Gupta et al.~\cite{gupta2018reducing} analyzed the historical ticket interactions between analysts and users, and proposed a system to requesting additional information when the user was raising a ticket based on the ticket description. Moreover, Shimpi et al.~\cite{shimpi2014problem} proposed an integrated framework to extracting issues from tickets under different scenarios, including system-generated and user-generated tickets. Han et al.~\cite{han2018towards} studied the problem of extracting software product names that are represented by various surface forms and informal abbreviations. They designed features to pricisely map abbreviations to the product and version by analyzing ticket language patterns. These approaches improved the efficiency of ticket analysis by extracting effective features of ticket descriptions. To fully exploit the information in the ticket data, our approach combines the description with the corresponding resolution to capture the effective and potential representation of tickets for ticket analysis.

\textbf{Ticket classification:} Potharaju et al.~\cite{potharaju2013juggling} proposed NetSieve which was a framework to infer problem symptoms, troubleshooting activities, and resolution actions based on analyzing the steps performed on each network entity from tickets. Zeng et al.~\cite{zeng2017knowledge} explored the hierarchical multi-label ticket classification problem and proposed GLabel, which was a greedy algorithm to search the optimal hierarchical multi-label for each ticket, where a novel contextual hierarchy loss was applied to maintain the accordance of labels assigned to each ticket. Zhou et al.~\cite{zhou2018multi} proposed an unsupervised ticket classification, which used the classic canonical analysis to extract ticket features for ticket labeling. Xu et al.~\cite{xu2018signature} constructed signatures to represent each type of ticket and proposed an algorithm to identifying the type of incoming tickets by comparing the similarity between the representation of incoming tickets and each type signature. To improve the effectiveness of measuring similarities among tickets, they also proposed a multi-view similarity measure framework that integrated the traditional similarity measures (e.g., Jaccard similarity,  NLCS)~\cite{xu2020multi}. These methods exploited semantic or syntactic analysis of ticket descriptions to improve the performance of ticket classification. However, in this paper, we considers the ticket automation as a Q\&A problem by translating the description into the corresponding resolution. Our proposed approach fully utilizes the sequential information to learn the representation of tickets which is capable of improving the performance of ticket classification.

\textbf{Ticket recommendation:} Ticket recommendation is mainly categorized as resolution recommendation and expert recommendation. Resolution recommendation is to provide appropriate historical ticket resolutions for the incoming ticket, while expert recommendation is to assign potential experts to address the incoming ticket. For resolution recommendation, Zhou et al.~\cite{zhou2016resolution} used the k-nearest neighbors algorithm to find out the historical tickets which are similar to the incoming ticket, and recommend the corresponding resolutions. However, the same exception would sometimes generate tickets with different descriptions due to the changes of system environment, which may weaken the effectiveness of the resolution recommendation. To solve this problem, they utilized structural corresponding learning based feature adaptation to maintain the coincidence of representations to these difference tickets ~\cite{zhou2015recommending}. Zhou et al.~\cite{zhou2017star} also used the siamese neural network to compare the similarity between historical tickets and the incoming one, and recommended the resolution which belonged to the most similar historical ticket. Wang et al.~\cite{wang2018aistar} cast resolution recommendation as a reinforcement learning problem and proposed an integrated framework based on hierarchical multi-armed bandits with the arm dependencies considered as a tree-structured hierarchy, while the framework recommends the resolution with the maximum reward at leaf node. For expert recommendation, Xu et al.~\cite{xu2018expert} proposed a two-stage expert recommendation model that if the initial expert cannot solve the ticket, the unsolved ticket will be transferred to the next expert. Furthermore, Han et al.~\cite{han2020uftr} proposed a unified ticket routing framework that incorporated four types of features (i.e., ticket, group, ticket-group, and group-group) for ranking the list of the expert recommendation. Recently, they also proposed a multi-view model to further improve the performance of the expert recommendation in their previous work, where the graph convolutional network was used to generate expert graph-view representation and the deep neural network was used to obtain the text-view feature~\cite{han2020deeprouting}. However, the traditional methods often ranked the expert and resolution recommendation list based on the similarity (e.g., the similarity between tickets, or the similarity between experts and tickets). In other words, if the first recommendation in the list fails, the second one will probably fail due to the incorrect representations of tickets or experts. In addition, these expert recommendation approaches relied on the ticket transfers among experts which emphasized the ability of experts while ignoring the root cause of tickets. The ticket recommendation is similar to the Q\&A task, which retrieves or generates the appropriate answer to the given question. Recently, Raheja et al.~\cite{dialogue} proposed a context-aware self-attention mechanism coupled with a hierarchical recurrent neural network to classify Dialogue Acts, effectively capturing utterance-level semantic text representations while maintaining high accuracy. In addition, Deng et al.~\cite{opinion-aware} proposed a unified model to exploit the opinion information from the reviews to facilitate the opinion-aware answer generation for a given product-related question, effectively generating opinionated and informative answers. These approaches performed well for the Q\&A task mainly comprising contextual natural language texts, however, they are ineffective for understanding ticket descriptions that are context-free and without polarity. To overcome these problems, SSR-TA is trained only based on the historical ticket descriptions and resolutions and is designed to recurrently generate the expert recommendation based on the assumption that the previous expert cannot solve the ticket.

\subsection{Sequence to Sequence}\label{subsec2}
Inspired by the recurrent neural network based encoder-decoder model, Sutskever et al.~\cite{sutskever2014sequence} proposed the seq2seq model, which is now widely used for many scenarios such as machine translation, image caption generation, and speech-to-text conversion. To improve the performance of seq2seq model, Bahdanau et al.~\cite{bahdanau2014neural} modified the original seq2seq model using the attention mechanism to reduce the information loss caused by compressing the source sentence into a hidden vector. Furthermore, Kenesh et al.~\cite{keneshloo2019deep} constructed the seq2seq model based on the reinforcement learning and tried to address the exposure bias and inconsistency between train/test measurement in the original seq2seq model. Pent et al.~\cite{peng2018sequence} proposed a seq2seq based approach for mapping natural language sentences to abstract meaning representation semantic graphs. Also, seq2seq was used in aspect term extraction, where the source sequence and target sequence is composed of words and labels ,respectively~\cite{ma2019exploring}. Huang et al.~\cite{huang2021multiplexed} proposed an end-to-end approach which can recognize multiple languages in images considering data imbalance between languages. Lewis et al.~\cite{bart} proposed a denoising autoencoder named BART, which combines Bidirectional and Auto-Regressive Transformers for pretraining sequence-to-sequence models, and BART is effective for text generation after fine-tuning. In addition, Mao et al.~\cite{gar} proposed GAR (Generation-Augmented Retrieval) for answering open-domain questions, which enriches the semantics of the queries by generating contexts for answer retrieval. These methods are effective for text generation and augmentation based on contextual information, however ticket descriptions are context-free, which is difficult for relevant information extraction. The principle of seq2seq model is to transform the sequential source data into the target data (e.g., English to Chinese, image to text, and question to answer). SSR-TA is to transform the ticket description into the corresponding resolution for capturing the potential and useful features which is capable of representing tickets and improve the performance of the expert recommendation.

\section{Method}\label{sec3}
\begin{figure}[ht] 
\centering 
\includegraphics[width=1\textwidth]{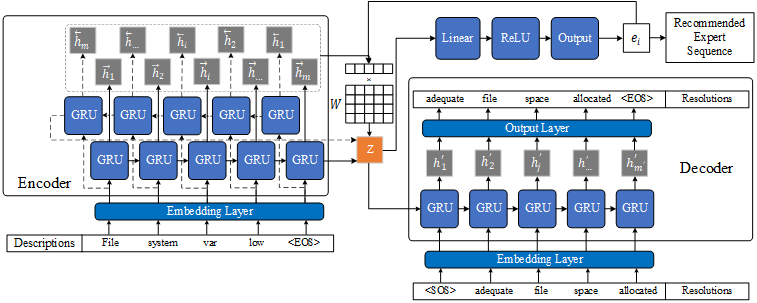} 
\caption{\centering The overview of SSR-TA} 
\label{fig.model} 
\end{figure}

SSR-TA mainly consists of three components: a description encoder, a resolution decoder, and an expert recommendation network. Fig.~\ref{fig.model} shows the overall structure of SSR-TA. The description encoder is used to transform the description of ticket into the potential representation of knowledge to the ticket (i.e., the hidden state). The resolution decoder, which helps the description encoder improve the effectiveness of representing the ticket, is to generate the corresponding resolution of the ticket based on the hidden state. The expert recommendation network will recurrently generate appropriate expert to solve the ticket based on the the hidden state which is fine-tuned using an attention mechanism.

\subsection{Data preprocessing}

\begin{table}[ht]
\begin{center}
\caption{A ticket example}\label{fig.ticketdata}
\begin{tabular}{@{}ll@{}}
\toprule
\makecell[r]{Fields} & Values\\
\midrule
\makecell[r]{ID} & 19027525 \\
\hline
\makecell[r]{Datetime} & 2014-04-29 06:00:12  \\
\hline
\makecell[r]{Type} & CONNECTIVITY  \\
\hline
\makecell[r]{Status} & CLOSED \\
\hline
\makecell[r]{Expert} & Lynn.Ridge \\
\hline
\makecell[r]{Description} & \makecell[l]{Failed to reconnect to PatrolAgent on host AVPMD623, port 3181. \\Will retry in 3 timer ticks.} \\
\hline
\makecell[r]{Resolution} & \makecell[l]{Job failed due to GFT connectivity issue. We have checked and \\restarted the job. It completed successfully.} \\
\botrule
\end{tabular}
\end{center}
\end{table}

Table~\ref{fig.ticketdata} illustrates the primary information of a ticket, mainly including structured and unstructured fields. The structured field contains ID, Type, Status, Datetime, and Expert, where these fields are generated automatically. In detail, ID is the index of the ticket, Type is the problem category, Status indicates whether the ticket is solved, Datetime is the generation time, and Expert is the expert who solved the ticket. The unstructured field contains the exception description generated by user/monitor system (i.e., Description) and the corresponding resolution proposed by the expert (i.e., Resolution). The notations mentioned in this paper are summarized in Table~\ref{tab.notation}.

\begin{table}[ht]
\begin{center}
\caption{Notations mentioned in this paper}\label{tab.notation}
\begin{tabular}{@{}ll@{}}
\toprule
Notation & Description\\
\midrule
\makecell[c]{$D$} & the description of ticket \\
\makecell[c]{$R$} & the resolution of ticket \\
\makecell[c]{$E$} & the expert label sequence of ticket \\
\makecell[c]{$m$} & the length of the description sequence \\
\makecell[c]{$m'$} & the length of the resolution sequence \\
\makecell[c]{$<EOS>$} & the special end-of-sentence symbol \\
\makecell[c]{$k$} & the $k$th expert in the sequence \\
\makecell[c]{$K$} & the number of experts \\
\makecell[c]{$\overline{E}$} & the recommended expert sequence of ticket \\
\makecell[c]{$h_i$} & the $i$th hidden state output by encoder \\
\makecell[c]{$z$} & the hidden state output by encoder last layer \\
\makecell[c]{$H$} & all hidden states output by encoder \\
\makecell[c]{$\tilde{W}$} & the attention parameter of the recommendation network  \\
\makecell[c]{$N$} & the number of experts recommended by model for a ticket \\
\makecell[c]{$p_k$} & the probability of the $k$th expert being recommended by the model \\
\makecell[c]{$set()$} & the same number of experts in the recommended sequence \\
\botrule
\end{tabular}
\end{center}
\end{table}

Suppose each ticket can be represented by a triple $(\bm{D}, \bm{R}, \bm{E})$, where $\bm{D}$ is the description of the ticket, $\bm{R}$ is the corresponding resolution, and $\bm{E}$ is the expert who solved the ticket. In detail, $\bm{D}$ can be represented by a sequence $[d_1, d_2, ..., d_i, ..., d_m, <EOS>]$, where $d_i$ means the $i_{th}$ word in the description and `$<EOS>$' is a special end-of-sentence symbol. Similarly, $\bm{R}$ is denoted as $[r_1, r_2, ..., r_j, ..., r_{m'}, <EOS>]$, where $r_{j}$ means the $j_{th}$ word in the resolution. $m$ is the length of the description sequence and $m'$ is the length of the resolution sequence. Each word is represented by a vector, which can be learned through the embedding layer based on the word embedding mechanism. Also, $\bm{E} = [e_1, e_2, ..., e_k, ..., e_K]$ is the expert label of the ticket, where $e_k=1$ indicates that the $k_{th}$ expert can solve the ticket and $e_k=0$ means the expert cannnot solve the ticket.

\subsection{Model Architecture}
 
\textbf{Description encoder:} We use a bidirectional Gate Recurrent Unit (GRU) network to construct the description encoder, and consider the word embedding of the ticket description $\bm{D}$ as the input. The output of the encoder $h_m$ is considered as the hidden state vector $\bm{z}$, which contains the contextual information of the description. At each step $i$, the corresponding hidden states $\overleftarrow{h}_i$ and $\overrightarrow{h}_i$ is generated based on the corresponding input $d_i$, the previous hidden states $\overleftarrow{h}_{i-1}$ and $\overrightarrow{h}_{i-1}$ from two directions, respectively:

\begin{equation}
    \overleftarrow{h}_{i}=f(d_{i} , \overleftarrow{h}_{i-1} ), \label{equ.encoder.left}
\end{equation}
\begin{equation}
    \overrightarrow{h}_{i}=f(d_{i} , \overrightarrow{h}_{i-1} ). \label{equ.encoder.right}
\end{equation}
The output of the encoder $h_m$ is generated as below:
\begin{equation}
    h_m = [\overleftarrow{h}_m, \overrightarrow{h}_m ]. \label{equ.encoder}
\end{equation}

\textbf{Resolution decoder:} The resolution decoder of SSR-TA is an original GRU that is trained to generate the resolution sequence $\textbf{R}$ based on the hidden state vector $\bm{z}$. In detail, the decoder generates the current hidden state $h_{j}^{'} $ and the prediction of current word $r_j$ based on the previous hidden state $h_{j-1}^{'} $ and the previous word $r_{j-1}$ at each step $j$, where the hidden state vector $\bm{z}$ is considered as the initial hidden state $h_{1}^{'} $. The conditional probability of resolution sequence is defined as:
\begin{equation}
    P(r_1, r_2, ..., r_{m'} \vert d_1, d_2, ..., d_m)=\prod \limits_{j=1}^n P(r_j\vert z,r_1, r_2, ..., r_{j-1}). \label{equ.decoder}
\end{equation}

\textbf{Expert recommendation:} The expert recommendation network will generate a sequence of experts recurrently based on the hidden state vector $\bm{z}$ from the description encoder. SSR-TA is designed to generate the next expert based on the assumption that the previous expert cannot solve the ticket. We adopt an attention mechanism to implement this recurrent recommendation process, defined as Equation~\ref{equ.attention1} and~\ref{equ.attention2}:
\begin{equation} 
    \bm{W}_n = tanh\left ( \tilde{W} \times \left [ \overline{E}_{n-1}, H \right ]  \right ){,} \label{equ.attention1}
\end{equation}
\begin{equation} 
    \bm{z}_n = \bm{W}_n \times H{,} \label{equ.attention2}
\end{equation}
where $\bm{W}_n$ is the attention weight for the $n_{th}$ recommendation, $\overline{E}_{n-1}$ is the $(n-1)_{th}$ recommendation result, $H$ is all the hidden states of the description encoder, $\bm{z}_n$ is the $n$th hidden state vector, and the $\Tilde{W}$ is a trained parameter. For example, the first hidden state vector $\bm{z}_1$ is used to generate the first recommended expert $e_1$. The attention weight $\bm{W}_2$ is generated based on the current recommendation result $\overline{E}_1$ and all hidden states from the description encoder $H$. Then, the attention weight will be combined with all the hidden state states $H$ to obtain the hidden state vector $\bm{z}_2$, which is used to recommend another appropriate expert $e_2$. Likewise, the subsequent recommendation results will be based on the assumption that the previous recommended expert failed to solve the ticket until the ticket is finally solved.

\subsection{Objective Function}
SSR-TA has three objectives, including the description translation, the accuracy of recommendation, and the disparity of recommendation. Therefore, three objective functions are designed to optimize our proposed model. The objective function for the description translation is to make the output of the resolution decoder as close to the real resolution as possible, which is defined as:
\begin{equation}
    Objective_{dt} = - \log_{}{P(\bm{R} \mid \bm{D})}{,} \label{equ.loss.seq}
\end{equation}
where the $\bm{D}$ and $\bm{R}$ are the ticket description and the corresponding resolution, respectively. 

For the recurrent recommendation network, the accuracy of recommendation means the probability of the expert recommended by SSR-TA each time is the true expert who solved the ticket, and the disparity of recommendation denotes the experts should be unique in the recommendation sequence. The objective function for the accuracy of recommendation is defined as Equation~\ref{equ.loss.acc}:
\begin{equation}
    Objective_{ar} = - \frac{1}{N}\sum_{n=1}^{N} \sum_{k=1}^{K} e_{k} \log_{}{(p_{k})}{,} \label{equ.loss.acc}
\end{equation}
where $e_k$ is the $k$th expert and $p_k$ denotes the probability of recommending the expert, $K$ is the number of experts, $N$ indicates the length of the recommended expert sequence. The objective function for the disparity of recommendation is defined as Equation~\ref{equ.loss.dis}:
\begin{equation}
    Objective_{dr} =  N-set(\overline{E} ){,}  \label{equ.loss.dis}
\end{equation}
where $\overline{E}$ is the recommended expert sequence, $set()$ represents the number of unique experts in the sequence.

Finally, three hyper-parameters $\alpha_1$, $\alpha_2$, and $\alpha_3$ are introduced to balance SSR-TA among the description translation, the accuracy of recommendation, and the disparity of recommendation. A joint objective function is defined as below:
\begin{equation}\label{equ.loss.all}
    Objective = - \alpha_1 \log_{}{P(R ,D )} - \alpha_2 \frac{1}{N}\sum_{n=1}^{N} \sum_{k=1}^{K} e_{k} \log_{}{(p_{k})} + \alpha_3  (N-set(\overline{E})){.} 
\end{equation}

\subsection{Train and Prediction}
The training process of SSR-TA is described in Algorithm~\ref{alg.rec}. The inputs of SSR-TA are the ticket description $\bm{D}$, the corresponding ticket resolution $\bm{R}$, the length of the recommended expert sequence $N$, and the expert label of the corresponding ticket $\bm{E}$. The output is the recommended expert sequence $S$. In detail, the model will generate an empty recommendation sequence $S$, and the first hidden state vector $\bm{z}_1$ is obtained from the description encoder. Then, based on $\bm{z}_1$, the resolution decoder predicts the corresponding resolution of the ticket, while the first recommendation result $\overline{E_1}$ is generated by the model, and the most probable expert $e_1$ in $\overline{E_1}$ is added to the sequence $S$. For the loop (i.e., Step 6 to 10), the model will generate the next hidden state vector $\bm{z}_n$ based on the previous recommendation result $\overline{E}_{n-1}$, and obtain the next recommendation expert $e_n$. This loop will be repeated until the recommendation sequence $S$ is filled by $N$ recommended experts. During the training process, the teach-forcing~\cite{DBLP:conf/nips/GoyalLZZCB16/teachforcing} is used to guarantee the efficient convergence while ensuring the robustness of the model. 

\begin{algorithm}[ht]
\caption{Seq2seq based Recurrent Expert Recommendation}\label{alg.rec}
\begin{algorithmic}[1]
\renewcommand{\algorithmicrequire}{\textbf{Input:}}  
\renewcommand{\algorithmicensure}{\textbf{Output:}} 
\Require
a description of the current ticket $D$; 
the resolution of the current ticket $R$; 
the length of recommended expert sequence $N$; 
the expert label of the ticket $E$.
\Ensure
the recommended expert sequence $S$.
\State S = $\emptyset$  //generate a empty sequence    
\State $z_1 = encoder(D)$  
\State $\overline{R} = decoder(z_1)$   //$\overline{R}$ is predicted resolution by decoder
\State $\overline{E_1} \Leftarrow recommendation(z_1)$  // get the first recommendation result
\State $e_1 \Leftarrow get(\overline{E_1})$  // get the most probable expert
\State $S.append(e_1)$
\While{$S.length < N$} // get the remaining recommended experts
    \State $\bm{W}_n = tanh\left ( \tilde{W} \times \left [ \overline{E}_{n-1}, H \right ]  \right )$
    \State $\bm{z}_n = \bm{W}_n \times H$  //get new hidden state vector
    \State $\overline{E_n} \Leftarrow recommendation(z_n)$  // get the $n_{th}$ recommendation result
    \State $e_n \Leftarrow get(\overline{E_n})$  // get the most probable expert 
    \State $S.append(e_n)$
\EndWhile
\end{algorithmic}
\end{algorithm}

The trained model can be used to recommend experts for incoming tickets. The process of prediction is similar to the training process, but only the ticket description will be used to obtain the recommended expert sequence. If the incoming ticket can be solved by an expert in the recommendation sequence, we consider that the recommendation is effective. Furthermore, the true resolver is ranked higher in the recommendation sequence, which means the model is more efficient. After the prediction, the resolution proposed by the true resolver for the corresponding ticket can be used to improve the robustness of the model. In other words, if the description translation loss of new tickets reaches the predefined threshold, the model can be updated based on these tickets.

\section{Experiments}\label{sec4}
In this section, we conduct experiments on two real-world datasets to evaluate the performance of SSR-TA for the expert recommendation, and mainly discuss the following issues:

\begin{enumerate}
    \item \textbf{Component combination}: Whether the resolution decoder and recurrent recommendation network are effective for the expert recommendations?
    
    \item \textbf{Seq2seq exploration}: Do the different structures of the seq2seq model have impacts on the performance of the expert recommendations?
    
    \item \textbf{Performance of expert recommendation}: How does the performance of SSR-TA compare to the performance of baselines?
    
    \item \textbf{Case study}: What is the difference between the effectiveness of SSR-TA and baselines on the real-world datasets?
\end{enumerate}

The implementations of SSR-TA and baselines are presented at Github: \href{https://github.com/coderxor/SSR-TA}{code}

\subsection{Datasets}

\begin{table}[ht]
\centering
\caption{The number of tickets and experts on two datasets}
\label{tab.ticket.number}
\begin{tabular}{ccccc}
\toprule
Dataset & All Ticket & Distinct Description & Distinct Resolution & Expert  \\
\midrule
TDS-a & 72977 & 15188 & 8574 & 64 \\
TDS-b & 23935 & 10693 & 7452  & 256\\
\botrule
\end{tabular}
\end{table}

The ticket datasets, TDS-a and TDS-b, were collected from different IBM IT service system. Table~\ref{tab.ticket.number} shows the number of tickets and experts in the corresponding datasets$\footnote{The datasets analysed during the current study are not publicly available due it involves private information from IBM but are available from the corresponding author on reasonable request.}$. Both datasets contain duplicate tickets and unsolved tickets, 15188 and 10693 tickets over two datasets were selected for experiments, and the experts solved tickets less than 50 are removed. Because almost all ticket descriptions contain numbers, dates, and special symbols (such as `WVPMA584 $\mid$ 03/02/2014'), and the description of the ticket submitted by the user may contain abbreviations and spelling errors (such as `immed' means immediate ). Then we performed symbol filtering, stop word removal, and lemmatization for description and resolution for all selected tickets. Some selected ticket examples from two datasets are shown in Table~\ref{tab.ticket.example}.

\begin{table*}[!ht]
    \scriptsize
            \caption{Some ticket examples from two datasets}
            \label{tab.ticket.example}
            \centering
            \renewcommand{\arraystretch}{1.3}
            \begin{tabular}{m{1cm}<{\centering} | m{2in} | m{2in}}
              \hline
              Dataset & Description & Resolution \\
              \hline
              \multirow{16}{*}{TDS-a}
                      & sppwa928: Available Real Memory is low. The percentage of available real memory is 5.0 percent. & ProblemSolutionText: we have validated the application and its BAU. \\
                      \cline{2-3}
                      & AVPMD504: File system / is low. The percentage of used space in the file system is 90 percent. Threshold: 90 percent. & Removed excess files and current filesystem usage now lower than threshold. According to System Operations Procedures closing the ticket accordingly. \\
                      \cline{2-3}
                      & lvpmi027: A high percentage of system CPU is being used.(99.98 percent) & Server CPU usage was checked and is within normal parameters of operation. This incident will fall into the Defect Prevention Process. \\
                      \cline{2-3}
                      & Patrol Agent Offline : Failed to reconnect to PatrolAgent on host AVPMD580, port 3181. Will retry in 3 timer ticks. & verified connectivity. \\
                      \cline{2-3}
                      & lppww665: The syslogd process is not running. & No actions taken since the process is running as expectedThis incident will fall into the Defect Prevention Process. \\
              \hline
              \multirow{16}{*}{TDS-b}
                       & b1pavreconapp01(161.178.193.234) is unreachable. The host has failed to respond to the ping request. & Server was rebooted under change \# CH165505. \\ 
                       \cline{2-3}
                       & 01.11 ET Job DOSW05P Abended at step DOS28090 with CC=SB37. & Insufficient space in output dataset. \\ 
                       \cline{2-3}
                       & Corrective Action ACR\_AC\_SVC\_WIN (101) failed. (Received during suppression) Service in alert state. & Service will be started automatically whenever required. \\ 
                       \cline{2-3}
                       & 174312 9015.00 SEV-2 TRANSFER STEP: FAILED FILE: TRANSEAPPRD. EPTRN3 \#M06EC122355Z9Q. & File has been re-transmitted successfully. \\ 
                       \cline{2-3}
                       & PowerHA Event reported by AVPMD592 AVPMD592 Mar 1 02:19:22 AVPMD592 user:notice HACMP for AIX: EVENT COMPLETED:network\_up\_complete AVPMD592 net\_ether\_02 0. & issue being worked on imr 19027973. \\
              \hline
            \end{tabular}
          \end{table*}

\subsection{Metrics}
To evaluate the performance of the expert recommendation, three metrics are used: Resolution Rate (RR)~\cite{DBLP:conf/bpm/SunTYAC10/content}, Mean Step To Resolver (MSTR)~\cite{shao2008efficient}, and Mean Reciprocal Rank (MRR)~\cite{han2020deeprouting}. 

RR represents the proportion of the tickets that can be solved. For example, if the true resolver (i.e., expert) to a ticket can be found in the recommendation sequence within the length $N$, it means the ticket can be solved. RR is defined as follows:
\begin{equation}
    RR(T)@N=\frac{\sum_{t\in T}^{} R(t)}{\mid T \mid}{,} \label{equ7}
\end{equation}
where $R(t)$ is $1$ if the ticket $t$ is solved and $0$ otherwise, while $T$ is the set of tickets.

MSTR represents the average number of steps to reach the resolver in the recommendation sequence. The smaller MSTR is, the more efficient the system is. The minimum value of MSTR is $1$, and MSTR is defined as:
\begin{equation}
    MSTR(T)=\frac{\sum_{t\in T}^{}P(t)}{\mid T^\ast \mid}{,} \label{equ6}
\end{equation}
where $T^\ast$ represents the set of tickets solved, and $\mid T^\ast \mid$ is the number of tickets in the set. For each ticket $t$, $P(t)$ means the position of the true resolver in the recommendation sequence. For example, if the recommendation sequence is $S = [e_1, e_2, e_n, ..., e_N]$ and true resolver is $e_n$, then $P(t)=3$. 

MRR represents the ranking of the true resolver corresponding to the ticket within the recommendation sequence:
\begin{equation}
    MRR(T)=\frac{1}{\mid T^\ast \mid}   {\textstyle \sum_{t=1}^{\mid T^\ast \mid} \frac{1}{rank_t} }{,} \label{equ8}
\end{equation}
where $rank_t$ denotes the position of the corresponding true resolver in the recommendation sequence for the ticket $t$.

\subsection{Baselines}
To evaluate the effectiveness of SSR-TA, we compare SSR-TA with several baselines.

\begin{enumerate}
    \item \textbf{SVM}~\cite{DBLP:journals/tist/ChangL11/LIBSVM}. SVM is a classic linear classifier, which is used to predict the probability of an expert solving a ticket. TF-IDF is used to obtain the representation vector of the ticket description which is considered as the input of SVM.
    
    \item \textbf{XGBoost}~\cite{DBLP:conf/kdd/ChenG16/XGBoost}. XGBoost is a scalable machine learning system for tree boosting, which is to predict whether the expert can solve the ticket. TF-IDF is still used to learn the representation of ticket descriptions.
    
    \item \textbf{RNN-uniform}~\cite{DBLP:conf/ijcai/LiuQH16/RNN}. RNN based text classification is a deep learning approach, and the LSTM layer can capture semantic information in variable-length sequences. The word embedding of ticket description is considered as the input of the RNN model which is to provide the expert recommendation sequence.

    \item \textbf{CNN-rand}~\cite{DBLP:conf/emnlp/Kim14/CNN}. This model casts the words in a ticket description as an embedding matrix, where the matrix will be classified through convolutional and max-pooling layers.

    \item \textbf{DeepRouting}~\cite{han2020deeprouting}. DeepRouting applies the text and graph similarity to predict the true resolver of the ticket. The deep structured semantic model is used to obtain the text-view representation of a ticket, and the graph convolutional network is adopted to learn the graph-view representation of expert relationships. For each expert, we sample 40 tickets it solved for training the GCN model. Moreover, for each positive pair (i.e., a ticket which is not solved by the expert), 19 negative experts are sampled.
    
    \item \textbf{BART}~\cite{bart}. BART is a denoising autoencoder combining Bidirectional and Auto-Regressive Transformers for pretraining sequence-to-sequence model. We applied pre-trained BART to recommend the ticket expert sequence by giving the ticket description.
    
\end{enumerate}

\subsection{Component combination}
In this section, we investigate whether the resolution decoder and the recurrent recommendation network are effective for the expert recommendation.

\begin{figure}[ht]
\centering
\subfigure[TDS-a]{\includegraphics[width=8cm]{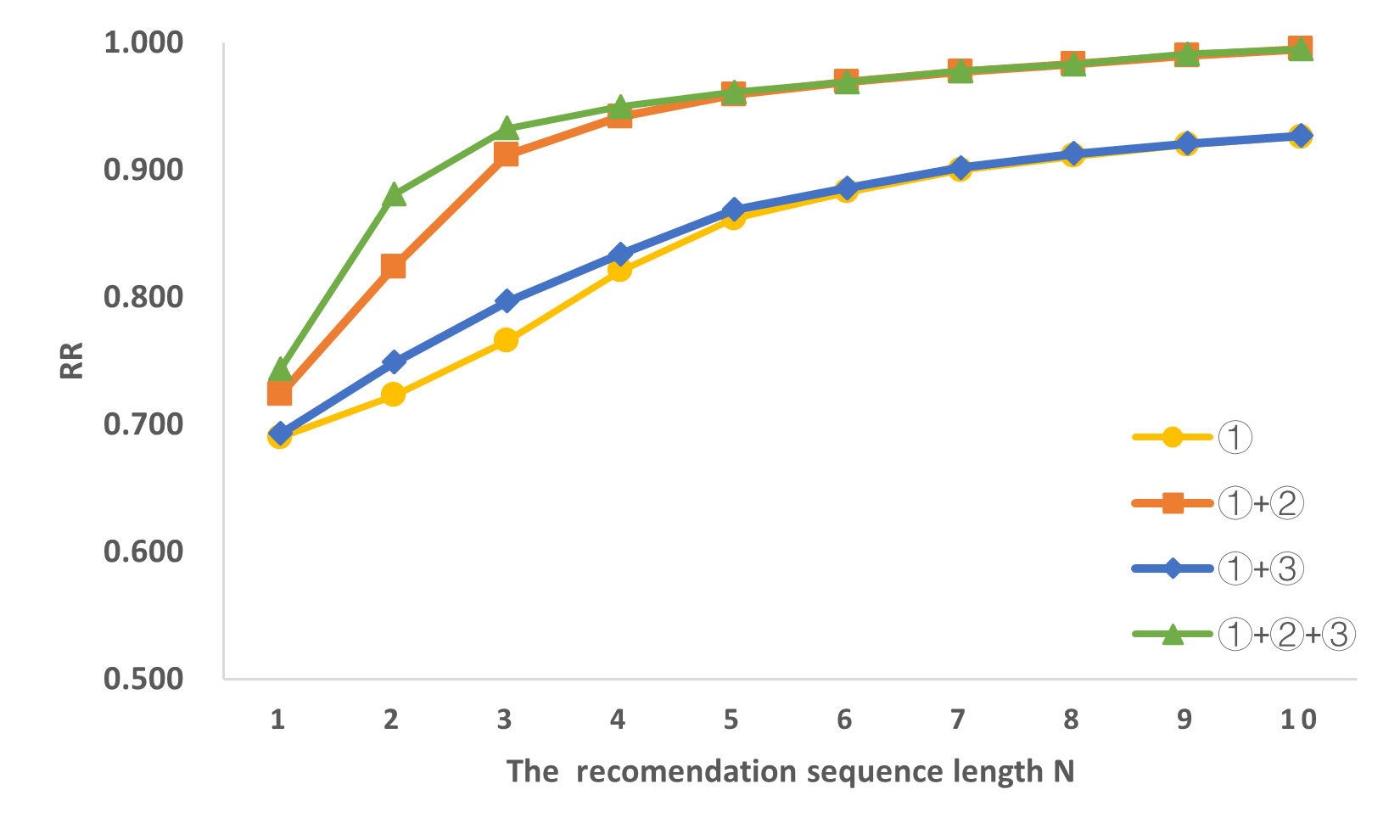}}
\subfigure[TDS-b]{\includegraphics[width=8cm]{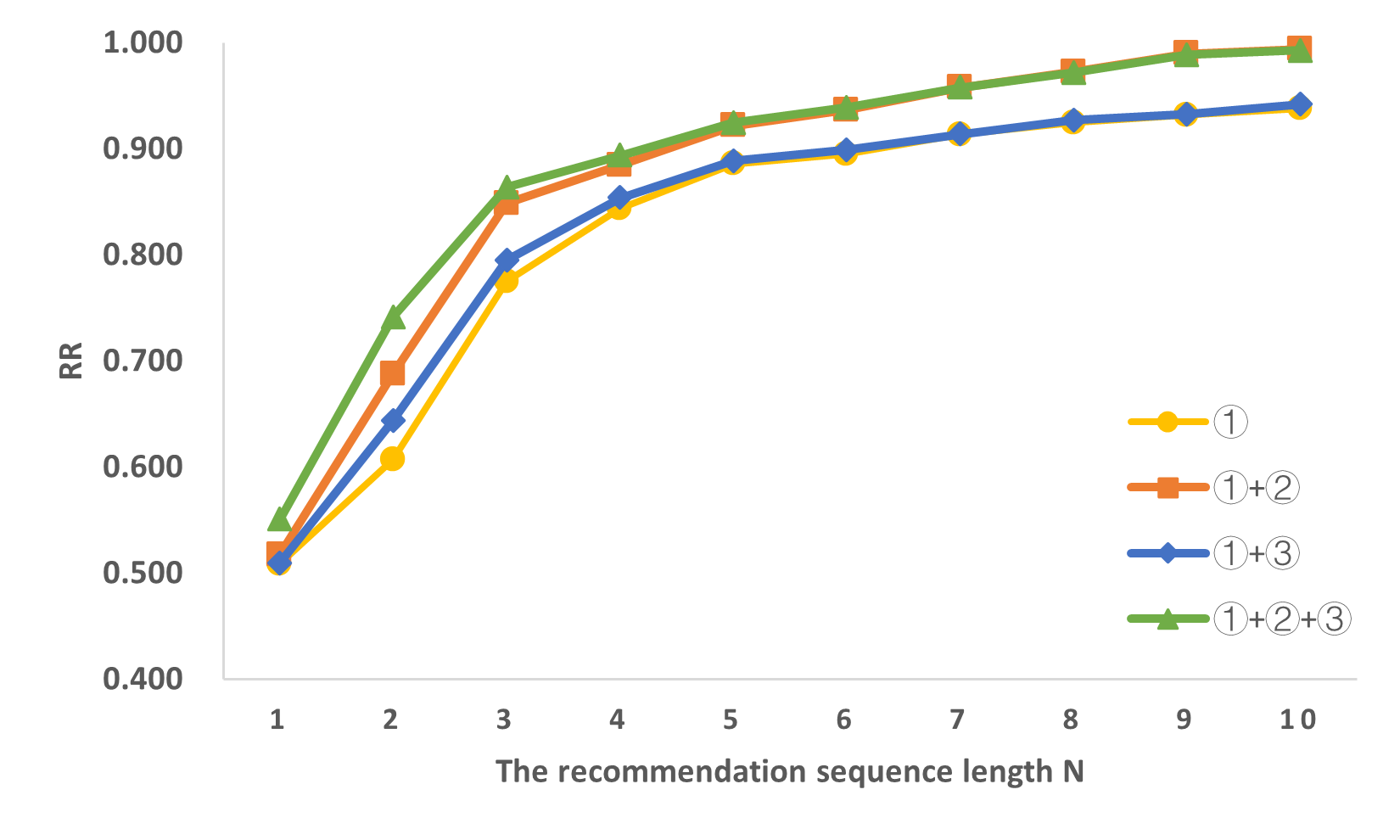}}
\caption{\centering The RR of component combination}
\label{fig.cc.rr}
\end{figure}

Fig.~\ref{fig.cc.rr} shows the resolution rate and Table~\ref{tab.diff-rrn.mstr} shows MRR and MSTR of SSR-TA of SSR-TA using different combinations of components, where \ding{172} is the description encoder, \ding{173} is the resolution decoder, and \ding{174} is the recurrent recommendation network. As observed from Fig.~\ref{fig.cc.rr} and Table~\ref{tab.diff-rrn.mstr}, the models without the resolution decoder (i.e., \ding{172} and \ding{172}+\ding{174}) perform worse than the seq2seq-based models (i.e., \ding{172}+\ding{173} and \ding{172}+\ding{173}+\ding{174}). In other words, the description translation loss is capable of improving the representation of ticket description which is effective for the expert recommendation. Moreover, the models with \ding{174} outperform those without \ding{174} (especially when the recommended sequence length is less than 5), which means that the recurrent recommendation network improves the ranking of the true resolver in the recommendation sequence. The phenomenon further supports our assumption that the model will recommend another appropriate expert when the previous experts in the sequence cannot solve the ticket. In summary, the experimental results show that the seq2seq structure improves the effectiveness of the expert recommendation, while the recurrent recommendation network improves the efficiency by reducing the number of steps to find true experts.

\begin{table}[ht]
\begin{center}
\begin{minipage}{\textwidth}
\caption{The MRR and MSTR of component combination}\label{tab.diff-rrn.mstr}
\begin{tabular*}{\textwidth}{@{\extracolsep{\fill}}lcccccc@{\extracolsep{\fill}}}
\toprule%
& \multicolumn{2}{@{}c@{}}{TDS-a} & \multicolumn{2}{@{}c@{}}{TDS-b} 
\\\cmidrule{2-3}\cmidrule{4-5}%
Model & MRR & MSTR & MRR & MSTR \\
\midrule
\ding{172} & 0.719 & 1.912 & 0.670 & 2.464 \\
\ding{172}+\ding{173} & 0.828 & 1.474 & 0.730 & 2.117 \\
\ding{172}+\ding{174} & 0.731 & 1.899 & 0.681 & 2.422 \\
\ding{172}+\ding{173}+\ding{174} & \textbf{0.843} & \textbf{1.352} & \textbf{0.743} & \textbf{2.043}\\
\botrule
\end{tabular*}
\footnotetext{The representations of terms are: \ding{172} the description encoder, \ding{173} the resolution decoder, \ding{174} the recurrent recommendation network}
\end{minipage}
\end{center}
\end{table}

\subsection{Seq2seq exploration}
In this section, we explore if the different structures of the seq2seq model have impacts on the performance for the expert recommendations. 

\begin{table}[ht]
\begin{center}
\begin{minipage}{\textwidth}
\caption{The MRR and MSTR of seq2seq exploration}\label{tab.diff-seq.mstr}
\begin{tabular*}{\textwidth}{@{\extracolsep{\fill}}lcccccc@{\extracolsep{\fill}}}
\toprule%
& \multicolumn{2}{@{}c@{}}{TDS-a} & \multicolumn{2}{@{}c@{}}{TDS-b} 
\\\cmidrule{2-3}\cmidrule{4-5}%
Model & MRR & MSTR & MRR & MSTR \\
\midrule
Bi-GRU & \textbf{0.843} & \textbf{1.352} & \textbf{0.743} & \textbf{2.043} \\
Bi-LSTM & 0.842 & \textbf{1.352} & 0.741 &	2.044\\
CNN & 0.815 & 1.619 & 0.691 & 2.379 \\
\botrule
\end{tabular*}
\end{minipage}
\end{center}
\end{table}

\begin{figure}[ht]
\centering
\subfigure[TDS-a]{\includegraphics[width=8cm]{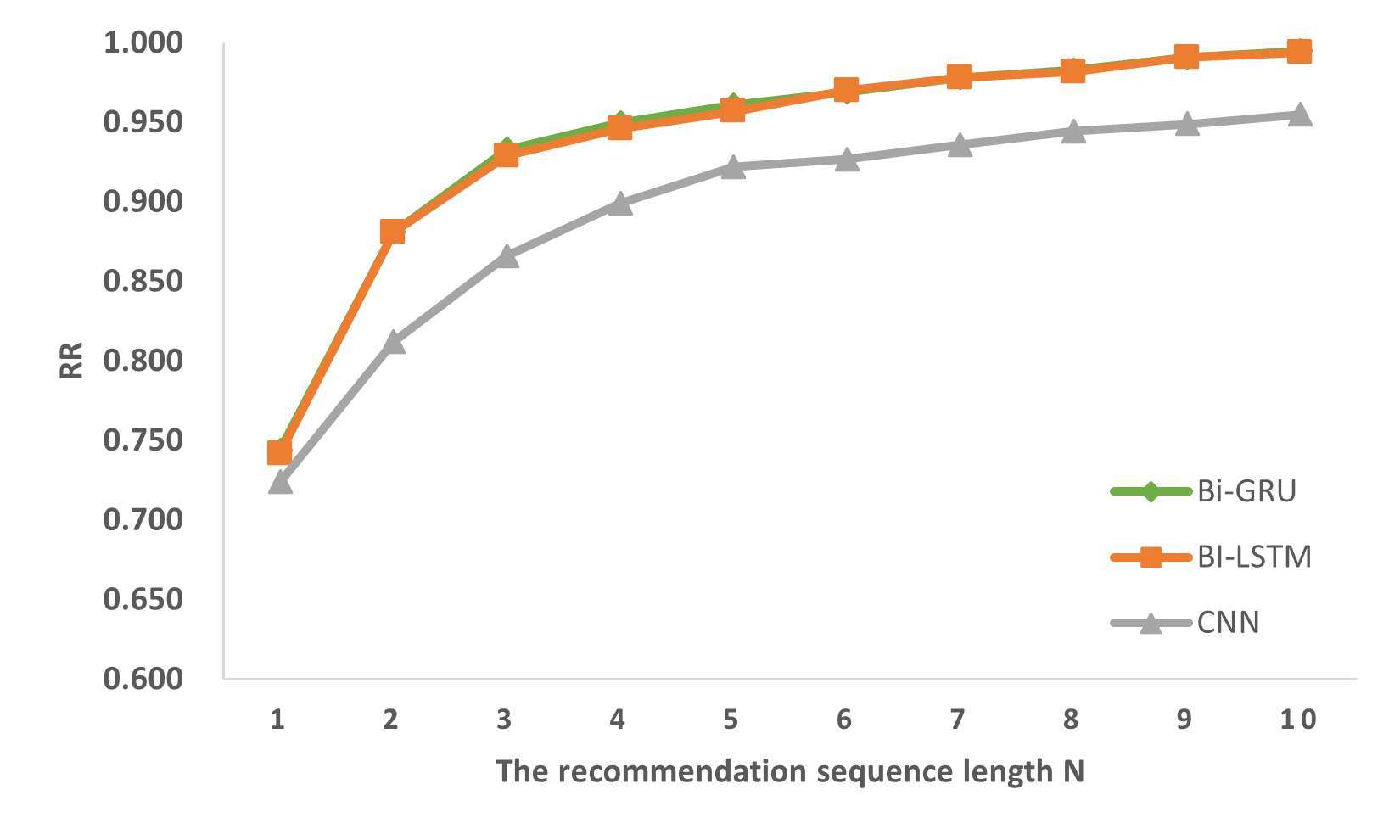}}
\subfigure[TDS-b]{\includegraphics[width=8cm]{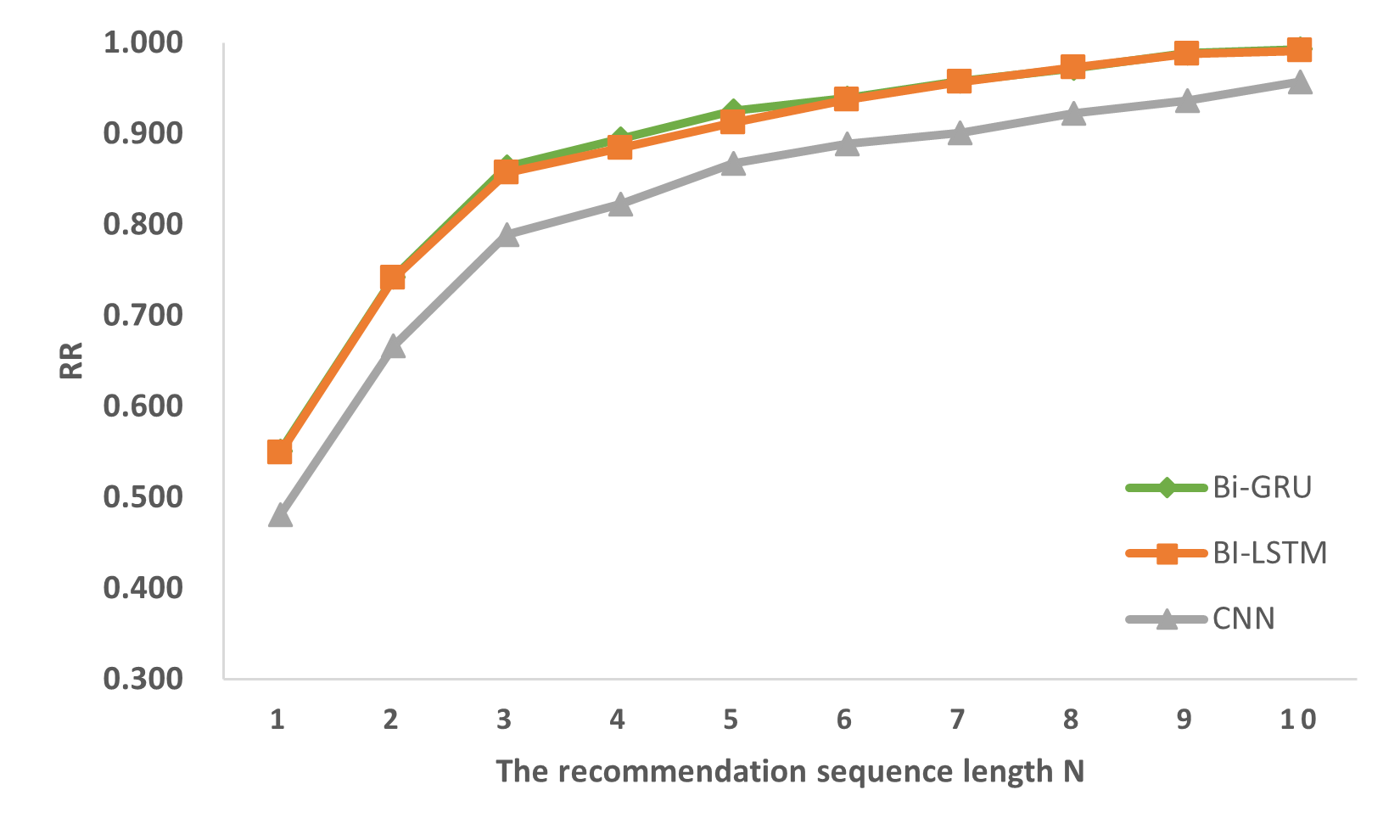}}
\caption{\centering The Resolution Rate of seq2seq exploration}
\label{fig.diff-seq.rr}  
\end{figure}

Fig.~\ref{fig.diff-seq.rr} represents the resolution rate and Table~\ref{tab.diff-seq.mstr} shows MRR and MSTR of SST-TA with different seq2seq structures, where Bi-GRU and Bi-LSTM are RNN based seq2seq models, and CNN based seq2seq model is composed of convolutional layers. The result shows that the RNN based models perform better than CNN based model (especially on TDS-a), which means the sequential features captured by the description translation are effective for the expert recommendation. In addition, the performance of RNN based models is close due to the similar structure of Bi-GRU and Bi-LSTM, while Bi-GRU is slightly better than Bi-LSTM. Overall, Bi-GRU is applied to construct SSR-TA because Bi-GRU has better performance and fewer parameters for faster training.

\subsection{Performance of expert recommendation}
\begin{figure}[ht]
\centering
\subfigure[TDS-a]{\includegraphics[width=8cm]{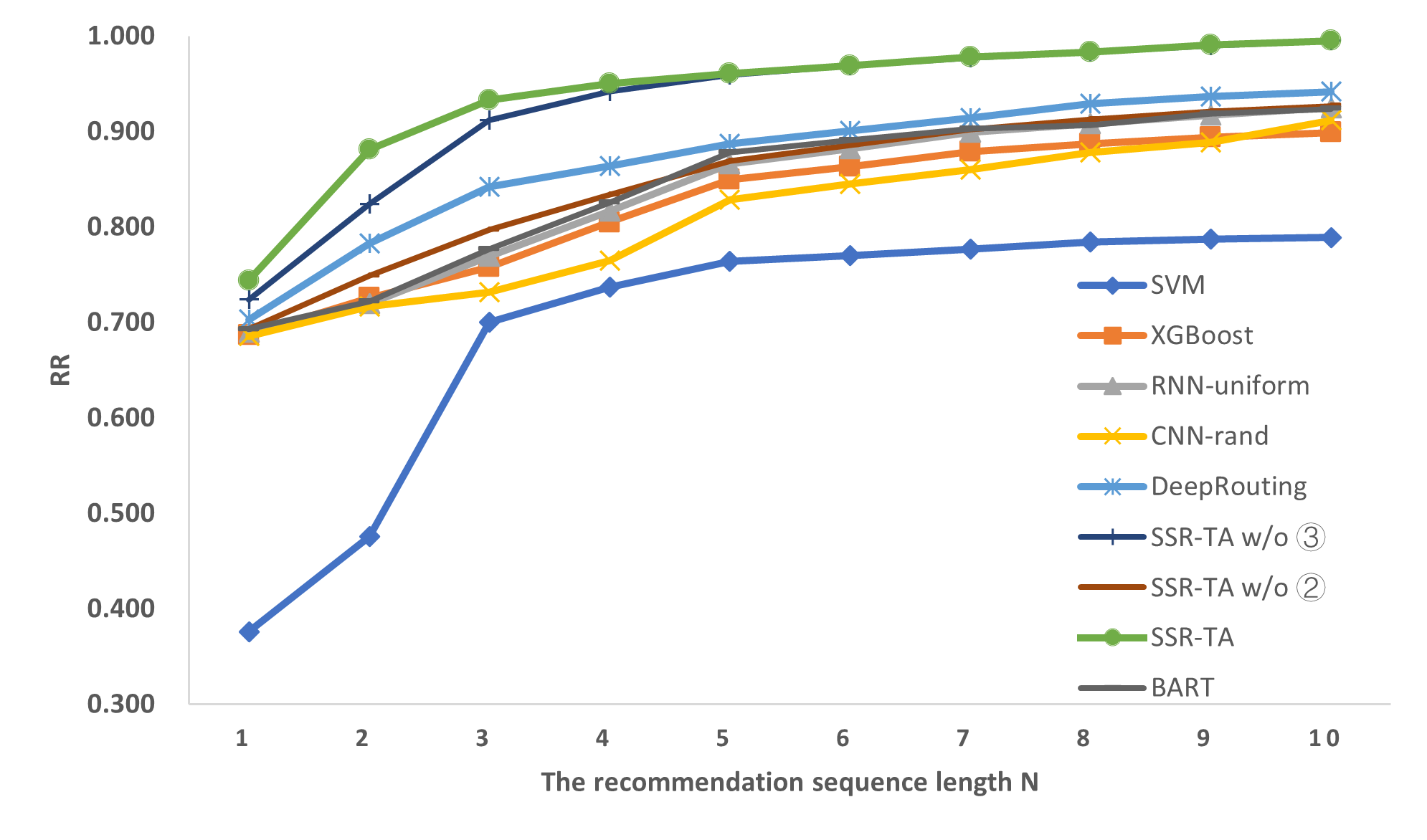}\label{fig.base-a}}
\subfigure[TDS-b]{\includegraphics[width=8cm]{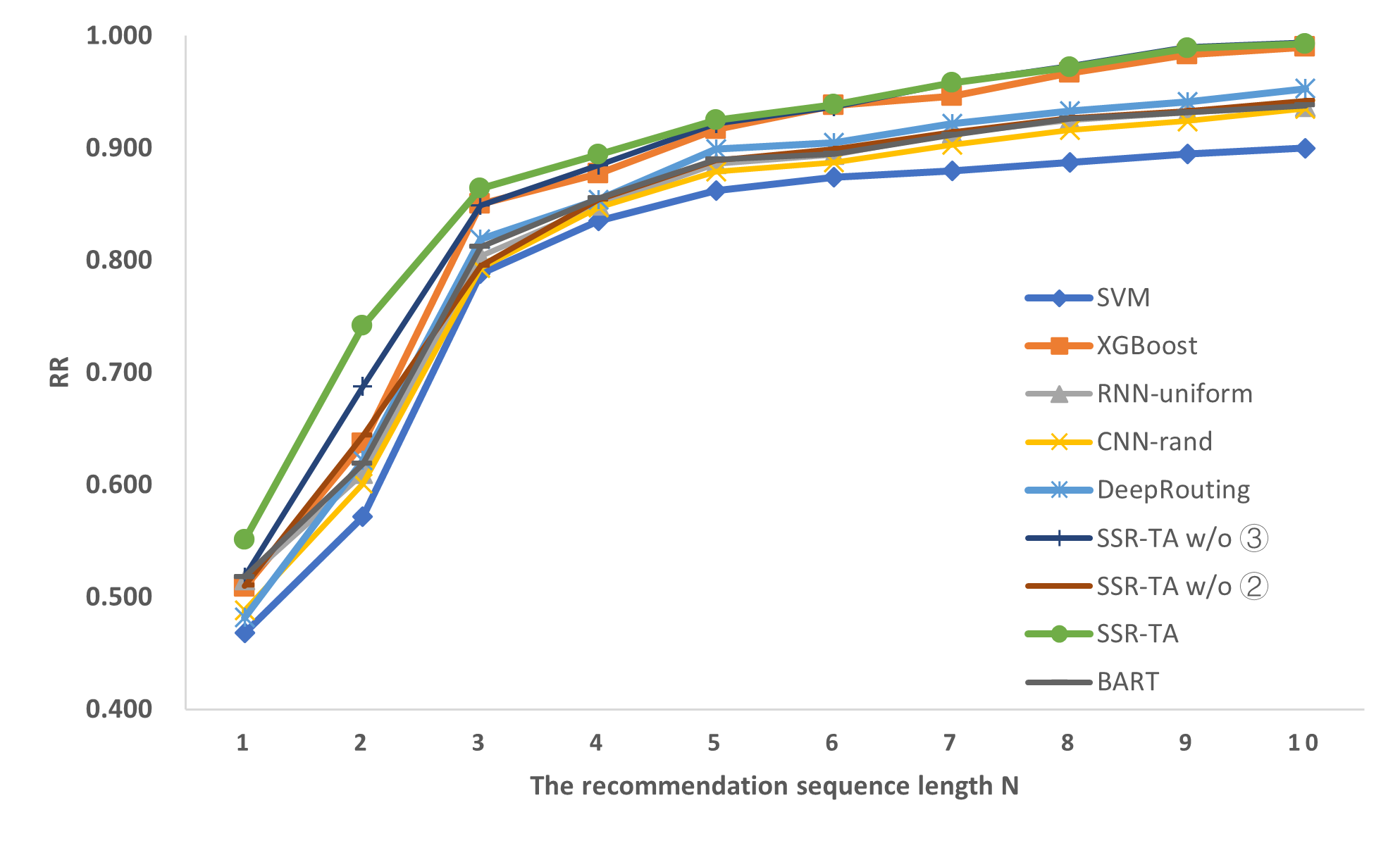}\label{fig.base-b}}
\caption{\centering The RR comparisons of SSR-TA with baselines}
\label{fig.baseline}
\end{figure}

In this section, we compare SSR-TA with the baselines on the read-world datasets, where Fig.~\ref{fig.baseline} illustrates the resolution rate and Table~\ref{tab.baseline} shows MRR and MSTR.

\begin{table}[ht]
\begin{center}
\begin{minipage}{\textwidth}
\caption{The MSTR and MRR comparisons of SSR-TA with baselines}\label{tab.baseline}
\begin{tabular*}{\textwidth}{@{\extracolsep{\fill}}lcccccc@{\extracolsep{\fill}}}
\toprule%
& \multicolumn{2}{@{}c@{}}{TDS-a} & \multicolumn{2}{@{}c@{}}{TDS-b} 
\\\cmidrule{2-3}\cmidrule{4-5}%
Model & MRR & MSTR & MRR & MSTR \\
\midrule
SVM & 0.575 & 2.788 &	0.629 &	2.614\\ 
XGBoost & 0.704 &	1.972 &	0.719 &	2.197\\ 
RNN-uniform & 0.720 & 1.947 & 0.676 & 2.451\\ 
CNN-rand & 0.692 &	2.123 &	0.669 &	2.485\\ 
DeepRouting & 0.767 & 1.793 &	0.686 &	2.401\\ 
BART & 0.723 & 1.923 & 0.678 & 2.450 \\
SSR-TA w/o \ding{174}  & 0.828 &	1.474 &	0.730 &	2.117 \\
SSR-TA w/o \ding{173}  & 0.731 &	1.899 &	0.681 &	2.422 \\
SSR-TA & \textbf{0.843} & \textbf{1.352} & \textbf{0.743} & \textbf{2.043}\\
\botrule
\end{tabular*}
\end{minipage}
\end{center}
\end{table}

As observed from Fig.~\ref{fig.baseline} and Table~\ref{tab.baseline}, SSR-TA outperforms other baselines. Notice that the RR of SSR-TA and DeepRouting increases faster than other baselines on TDS-a, especially when the length of the recommendation sequence is less than 3 (i.e., $N \le 3$). The experimental results show that the recurrent recommendation of SSR-TA and the ticket transfer of DeepRouting effectively improve the ranking of the true resolver in the recommendation sequence. In addition, on TDS-b, different from DeepRouting, SSR-TA also performs better in RR when $N \le 3$, which means that the generalization ability of the recurrent recommendation network is better than that of the ticket transfer. Additionally, the pre-trained BART is outperformed by DeepRouting and SSR-TA. Compared to BART, which only makes one-time recommendations based on ticket descriptions, SSR-TA adjusts the subsequent recommendations with the recurrent recommendation network, which reduces the steps of resolving tickets. Furthermore, the performance of SSR-TA w/o \ding{173} is close to that of other baselines, which further shows that the resolution decoder is necessary and effective for the expert recommendation. 

In addition, Table~\ref{tab.trainingtime} shows the training time of SSR-TA and baselines. All models use the same datasets(TDS-a and TDS-b) and experimental environment and are implemented with PyTorch, and the test system is Linux with 32 GB of RAM and 1080 Ti GPU. The results show that the training time of SSR-TA is more than that of traditional classification models (e.g., SVM and XGBoost), close to other RNN-based methods (e.g., RNN-uniform), but less than that of DeepRouting and BART. DeepRouting is time-consuming due to the application of the graph model GCN, while BART is constructed based on BERT, which is a model comprising millions of parameters. In summary, compared to DeepRouting and BART, the performance of SSR-TA is more effective and efficient, while comparing to the traditional methods (i.e., SVM and XGBoost) and neural network-based methods (i.e., RNN-uniform and CNN-rand), SSR-TA is more effective with appropriate efficiency.

In conclusion, the description translation (i.e., the description encoder and the resolution decoder) improves the effectiveness, while the recurrent recommendation network improves the efficiency (i.e., the ranking of the true resolver in the recommendation sequence) of the expert recommendation.

\begin{table}[ht]
\centering
\caption{The training time comparisons of SSR-TA with baselines}\label{tab.trainingtime}
\begin{tabular}{lll}
\toprule%
& \multicolumn{2}{l}{Training Time (seconds)} \\
\cmidrule{2-3}
Model & TDS-a & TDS-b \\
\midrule
SVM & 321 & 68 \\ 
XGBoost & 10 & 3 \\ 
RNN-uniform & 1635 & 685 \\ 
CNN-rand & 807 & 316 \\ 
DeepRouting & 5913 & 2581 \\ 
BART & 11843 & 5059 \\
SSR-TA w/o \ding{174}  & 1832 &	747 \\
SSR-TA w/o \ding{173}  & 1683 &	664 \\
SSR-TA & 1866 & 794 \\
\botrule
\end{tabular}
\end{table}

\subsection{Case study}
As observed from the comparison between SSR-TA and the baselines in the real-world datasets, we notice that the performance of the baselines and SSR-TA are similar on TDS-b, which is different from that on TDS-a. Therefore, we further investigate the difference between TDS-a and TDS-b in this section.

\begin{table}[ht]
\centering
\caption{The number of user-generated and system-generated tickets on two datasets}
\label{tab.ticket.distribution}
\begin{tabular}{llll}
\toprule
Dataset & User-generated & System-gengerated & Total  \\
\midrule
TDS-a   & \makecell[c]{13680}          & \makecell[c]{1508}              & 15188  \\
TDS-b   & \makecell[c]{4103}           & \makecell[c]{6590}              & 10693  \\
\botrule
\end{tabular}
\end{table}

As mentioned earlier, TDS-a and TDS-b are collected from different IT managed service systems, therefore, there exist some differences between tickets. For example, the description of a user-generated ticket ``AGPMA506: A high percentage of CPU is being used. (93 percent)'', which is similar to the natural language. However, the system automatically generated tickets such as ``EXPCHNG\_8193\_WVPMA582 WVPMA582 06:05:43 Message Error Information : 1) Exception Details : Error Code :8193 Severity Code :2 Message :Exception occurred at Business Layer while Adding SourceRequired Error Type'', which are mainly constructed by a template containing a large number of non-English words (e.g., EXPCHNG). In other words, the user-generated tickets full of sequential information can be effectively learned by the RNN-based model, while the RNN-based model is not capable of capturing the representation of the system-generated tickets, which is without the effective sequential information. Table~\ref{tab.ticket.distribution} presents the proportion of the user-generated tickets and the system-generated tickets in TDS-a and TDS-b. As observed from Table~\ref{tab.ticket.distribution}, the proportion of system-generated tickets in TDS-b is much higher than that in TDS-a. Moreover, Fig.~\subref{fig.base-a} shows that RNN-uniform performs better than other traditional top-N recommendation models, while RNN-uniform performs similar to other baselines in Fig.~\subref{fig.base-b}. In other words, the RNN based models (i.e., SSR-TA and RNN-uniform), which are capable of capturing the sequential feature, perform better on TDS-a containing more user-generated tickets.

\subsection{Discussion}
The proposed SSR-TA in this paper is used to recommend experts to solve tickets. The seq2seq model captures feature information by translating ticket descriptions into corresponding resolutions, and the recurrent recommendation network is designed to obtain expert recommendation sequences.

For ticket recommendation, SSR-TA applied the seq2seq model to combine the features from ticket description and resolution to assist the expert recommendation, and it also can be used for ticket resolution generation. Moreover, the recurrent recommendation network adopts an attention mechanism to generate the expert recommendation sequences based on the assumption that the previous expert in the sequence failed to solve tickets. The recurrent recommendation network improves the ranking of the true resolver in the recommendation sequence, which means the model reduces the steps to find appropriate experts for solving tickets. In other words, the application of SSR-TA is capable of solving anomalies more quickly and saving more human costs in practice. 

Finally, there exist some limitations to our proposed method. The performance of expert recommendations for the system-generated tickets is worse than that for the user-generated tickets. Because the seq2seq model (i.e., SSR-TA) is specialized in capturing the sequential information from natural language (i.e., user-generated tickets), while the system-generated tickets mainly consist of unfamiliar words and specific terms without contextual information. In addition, training SSR-TA is much more time-consuming than training those traditional machine learning methods (i.e., SVM and XGBoost), which is difficult to apply in practice. Furthermore, limited to the architecture of SSR-TA, SSR-TA can select and recommend experts or existing resolutions for the known tickets, however, SSR-TA cannot generate personalized resolutions for unknown tickets.

\section{Conclusion}\label{sec5}
In this paper, we proposed a seq2seq based model SSR-TA combined with a recurrent recommendation network to recommend appropriate experts for the ticket automation. The description of solved tickets is translated into the corresponding resolution for capturing the potential and useful features, which is capable of representing tickets and improving the effectiveness of the expert recommendation. In addition, the recurrent recommendation network improves the ranking of the true resolver in the recommendation sequence based on the assumption that the previous expert cannot solve the ticket. The experimental results show that SSR-TA improves the effectiveness and the efficiency of the expert recommendation. 

In future work, we will continue to study the following issues. First, we are going to improve the initial resolution rate of the model in processing system-generated tickets. Second, the decoder of SSR-TA is not used when recommending an expert sequence for an incoming ticket. We will recommend historical resolutions for the incoming ticket using the similarity between historical resolutions and the decoder output. Finally, we plan to explore more variants of seq2seq for ticket expert recommendation.

\bmhead{Acknowledgments}
This work was supported by the National Natural Science Foundation of China under Grant No. 61872186, the Natural Science Foundation of Nanjing University of Posts and Telecommunications (Grant No. NY221070).


\bibliography{Manuscript}


\end{document}